\documentclass[preprint]{revtex4-2}
\usepackage{graphicx}
\usepackage{dcolumn}
\usepackage{bm}

\begin{document}
	

\title{A complete, parallel and autonomous photonic neural network in a semiconductor multimode laser}

\author{Xavier Porte}
\affiliation{FEMTO-ST/Optics Dept., Univ. Bourgogne Franche-Comt\'{e},  UMR CNRS 6174, 15B avenue des Montboucons, 25030 Besan\c{c}on Cedex, France}

\author{Anas Skalli}
\affiliation{FEMTO-ST/Optics Dept., Univ. Bourgogne Franche-Comt\'{e},  UMR CNRS 6174, 15B avenue des Montboucons, 25030 Besan\c{c}on Cedex, France}

\author{Nasibeh Haghighi}
\affiliation{Institut für Festk\"orperphysik, Technische Universit\"at Berlin, Hardenbergstra\ss e 36, 10623 Berlin, Germany}

\author{Stephan Reitzenstein}
\affiliation{Institut für Festk\"orperphysik, Technische Universit\"at Berlin, Hardenbergstra\ss e 36, 10623 Berlin, Germany}

\author{James A. Lott}
\affiliation{Institut für Festk\"orperphysik, Technische Universit\"at Berlin, Hardenbergstra\ss e 36, 10623 Berlin, Germany}

\author{Daniel Brunner}
\affiliation{FEMTO-ST/Optics Dept., Univ. Bourgogne Franche-Comt\'{e}, UMR CNRS 6174, \\ 15B avenue des Montboucons, 25030 Besan\c{c}on Cedex, France,  e-mail: daniel.brunner@femto-st.fr}

\date{\today}

\begin{abstract}

Neural networks are one of the disruptive computing concepts of our time.
However, they fundamentally differ from classical, algorithmic computing in a number of fundamental aspects.
These differences result in equally fundamental, severe and relevant challenges for neural network computing using current computing substrates.
Neural networks urge for parallelism across the entire processor and for a co-location of memory and arithmetic, i.e. beyond von Neumann architectures.
Parallelism in particular made photonics a highly promising platform, yet until now scalable and integratable concepts are scarce.
Here, we demonstrate for the first time how a \emph{fully parallel} and \emph{fully implemented} photonic neural network can be realized using spatially distributed modes of an efficient and fast semiconductor laser.
Importantly, all neural network connections are realized in hardware, and our processor produces results without pre- or post-processing.
130+ nodes are implemented in a large-area vertical cavity surface emitting laser, input and output weights are realized via the complex transmission matrix of a multimode fiber and a digital micro-mirror array, respectively.
We train the readout weights to perform 2-bit header recognition, a 2-bit XOR and 2-bit digital analog conversion, and obtain $<0.9\cdot10^{-3}$ and $2.9\cdot10^{-2}$ error rates for digit recognition and XOR, respectively.
Finally, the digital analog conversion can be realized with a standard deviation of only $5.4\cdot10^{-2}$.
Our system is scalable to much larger sizes and to bandwidths in excess of 20 GHz. 

\end{abstract}

\maketitle

\section{Introduction}

Artificial neural networks (ANNs) were conceptualized already in the 40s of the previous century, with providing a ‘logical calculus of the ideas immanent in nervous activity’ as the original ambition \cite{McCulloch1943}.
In line with this spirit, nonlinear nodes standing in as neurons form the ANN with the help of connections, much like synapses, dendrites and axons connect biological neurons.
Such a \emph{parallel} architecture distributes computing across a usually large number of simple nonlinear nodes, and programming becomes modifying the network's topology in a step which often is refered to as training.
Besides several important breakthroughs such as error-back propagation to optimize, i.e. train an ANN \cite{Rumelhart1986}, it was the increasing availability of powerful computing hardware \cite{Raina2009} which turned ANNs into a technology of primary societal \cite{LeCun2015}, scientific \cite{Senior2020} and economic relevance.

High-performance computing hardware therefore is a must for ANNs.
The reason lies in the core principles of a network: the number of possible connections scales quadratically with the number of network nodes.
All links of an ANN's interconnect must be queried before a computational result is obtained, and network-based computing consequently poses plenty of hardware challenges.
Most importantly, it has so far been impossible to integrate parallel and large ANNs, and application specific circuits employed today still heavily rely on serial routing.
Integration of such parallel networks in electronic circuits is highly challenging due to: (i) the strong energy penalty for switching the large number of signaling wires; and (ii) a fundamental incompatibility with two-dimensional (2D) integration \cite{Lin2020}.

Photonics is a promising alternative as it mitigates much of the energy cost of information transduction across a network.
This potential advantage over electronics was realized already decades ago \cite{Farhat1985,Lohmann1990} and is detailed in \cite{Miller2017}.
Reducing energy deposition into the substrate could enable novel three-dimensional (3D) integration \cite{Moughames2020} to ultimately realize scalable integration of parallel ANNs \cite{Dinc2020}.
Optical ANNs did outperform their electronic counterparts already at a previous stage \cite{Li1993}, and their superior scaling has recently again been demonstrated using a large scale free-space experiment \cite{Rafayelyan2020}.
Finally, ANNs implemented in silicon photonics \cite{ Tait2014,Shen2016,Feldmann2019} promise ultra-high speed, ultra low latency and the benefits of integration leveraging the tools of silicon and 2D lithography.

However, the vast majority of large scale and parallel photonic ANN demonstrations are not standalone and autonomous.
Many such proof-of-concepts lack one or several of an ANN's constituents, i.e. neurons, connections or learning; or require substantial involvement of a classical electronic computer in the creation of the ANN's state, are not parallel or run on substrates too exotic for realistic integration in mid-term technological platforms.
Here, we address all these points and experimentally demonstrate a fully parallel photonic reservoir computer \cite{Jaeger2004,VanDerSande2017} with 131 neurons that we train and operate online and in realtime with minimal interference from a control personal computer.
As photonic neuron substrate we use the complex multimode field of an injection locked large area vertical cavity surface emitting laser (LA-VCSEL) of $\sim20~\mu$m diameter emitting around 920~nm.
Our LA-VCSEL follows design principles developed for high bandwidth and high efficiency VCSEL arrays \cite{Haghighi2021} and was fabricated by hand via standard ultraviolet contact mask lithography, dry (inductively coupled plasma-reactive ion etching) mesa etching, and thin-film metal evaporation and lift-off methods in a university cleanroom \cite{Haghighi2020,Haghighi2019}.
All the photonic ANN's connections are implemented in hardware.
The complex transfer matrix of a multimode fiber (mm-fiber) \cite{Carpenter2016} couples the LA-VCSEL to the injected information.
Intra-cavity fields and carrier diffusion intrinsic to LA-VCSELs recurrently couple the photonic neurons \cite{Mulet2002}, and trainable readout weights are encoded on a digital micro-mirror device (DMD) \cite{Bueno2016}.
Input information is Boolean-encoded on another DMD, and the photonic ANN's output is photo-detected to directly provide the computational result.

We operate our recurrent photonic ANN in its steady state, and its bandwidth is limited by the input DMD's frame rate to around 100 inferences per second.
However, such semiconductor lasers can be modulated extremely fast, often in excess of 20 GHz, using either electronic modulation \cite{Haghighi2020} or optical injection \cite{Simpson1995}, which therefore corresponds to the potential intrinsic speed of our system.
We train the readout weights to perform 2-bit header recognition, a 2-bit XOR and 2-bit digital analog conversion (DAC).
For the 2-bit header recognition our systems achieves an excellent error rate smaller than $0.9\cdot10^{-3}$, for the more demanding XOR it is still a respectable $2.9\cdot10^{-2}$.
Finally, the digital analog conversion can be realized with a standard deviation of only $5.4\cdot10^{-2}$.
In this context it is important to highlight the fully analog nature of our photonic ANN.

Our system overcomes several challenges of the field.
First, we obtain very low error rates in a fully parallel photonic ANN where \emph{each} of the system's constituents is hardware implemented.
Second, we exclusively utilize readily available technology and components with demonstrated reliability and cost effective manufacturing.
Third, the system is scalable in size to much larger networks in excess of 1000 neurons per layer through readily available LA-VCSELs with apertures exceeding $100~\mu$m.
And finally, our concept provides a clear and, most importantly, \emph{feasible} road-map for scaling to bandwidths in excess of 10 GHz.

\begin{figure}[t]
\centering
\includegraphics[width=1\columnwidth]{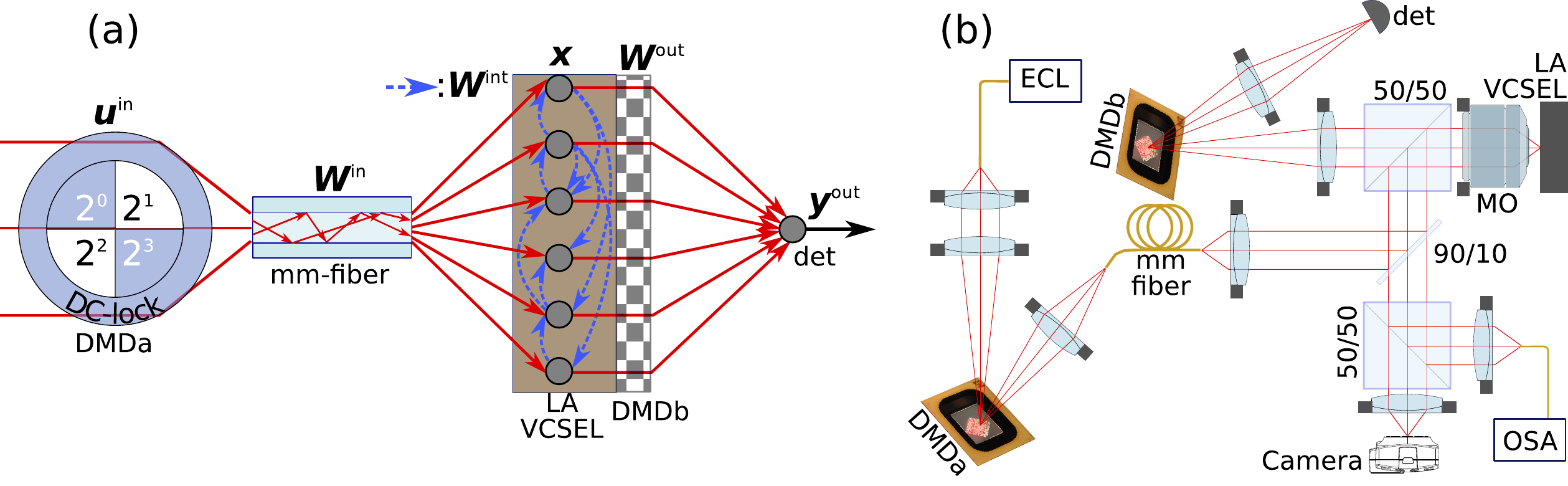}
\caption{ (a) Schematic illustration of the photonic ANN's sections linked to their corresponding physical devices.
	A digital micro-mirror device (DMD, DMDa) encodes intput information $\mathbf{u}^{\mathrm{in}}$, which is mixed through the complex transfer matrix of a multimode fiber (mm-fiber). The large-area vertical cavity surface emitting laser (LA-VCSEL) acts as recurrent reservoir with state $\mathbf{x}$, providing device-inherent internal coupling $\mathbf{W}^{\mathrm{int}}$.
	DMDb implements programmable Boolean readout weights $\mathbf{W}^{\mathrm{out}}$, and a detector records computational result $y^{\mathrm{out}}$.
	(b) Experimental arrangement of the corresponding components.
	Several beam splitters (50/50 and 90/10) guide the optical signals, and an infinity corrected MAG10 microscope objective (MO) images the injection onto the LA-VCSEL, and its emission onto DMDb and the camera.
	Emission spectra are recorded with an optical spectrum analyzer (OSA).}
\label{fig:ConceptExperiment}
\end{figure}

\section{\label{sec:Concept} A photonic ANN in a complex and continuous nonlinear media}

Parallelism in our photonic ANN is based on spatial multiplexing, and the correspondence between the ANN's conceptual sections and the relevant photonic hardware is schematically illustrated in Fig. \ref{fig:ConceptExperiment}(a).
The collimated beam of a tunable external cavity laser (ECL, Toptica CTL 950) illuminates an area of DMDa (Vialux XGA 0.7" V4100) with a Gaussian cross-section of 90 pixels in diameter.
A DMD comprises mirrors which can be switched between $\pm12^{\circ}$, and here mirrors in the -12$^{\circ}$ orientation inject information into the ANN.
Input information $\mathbf{u}^{\mathrm{in}}$ is therefore a Boolean matrix.
Within the illuminated area we define a ring with 10-pixels thickness that is constantly active.
This DC-locking signal $\mathbf{u}^{\mathrm{DC}}$ stabilises the LA-VCSEL's potentially autonomous free-running dynamics and hence allows consistency \cite{Bueno2016}.
Inside the remaining illuminated area we define $n$ sub-sections, much like wedges of a pie, with which we encode input information $\mathbf{u}^{\mathrm{d}}$ for digit $d$ in an $n$-bit binary representation.
In Fig. \ref{fig:ConceptExperiment}(a), input information $\mathbf{u}^{\mathrm{in}}=\mathbf{u}^{\mathrm{DC}}+\mathbf{u}^{\mathrm{d}}$ corresponds to injecting digit $d=9$ in a representation with $n=4$ bits.
The resulting, spatially-modulated beam is collected by a mm-fiber that implements complex input weights $\mathbf{W}^{\mathrm{in}}$ through optical mode mixing induced by birefringence and scattering \cite{Carpenter2016}.
Using the common estimation based on a fiber's $V$ number we approximate that our input fiber supports between 30 and 40 modes, i.e. an input matrix of equal size.

The near-field at the mm-fiber's output facet is injected into the LA-VCSEL.
Semiconductor lasers react highly nonlinearly to optical injection, and previously we showed that this nonlinear response can be harvested to implement the nonlinear nodes of an ANN \cite{Brunner2013a,Vatin2019}.
The feasibility of this concept was shown numerically for 33~GHz \cite{Estebanez2020}.
However, other than in \cite{Brunner2013a} here each neuron is physically implemented in parallel via the LA-VCSEL's complex and high-dimensional multimode field.
Furthermore, optical diffraction of the LA-VCSEL's intra-cavity field as well as carrier diffusion induce interactions between different locations, i.e. coupling between our photonic nodes \cite{Mulet2002}.
Dynamics induced by optical injection are governed by an equation of the type
\begin{equation}\label{eq:LAVCSEL}
\tau \dot{\mathbf{{x}}}(t) = -\mathbf{x}(t) + \mathbf{f}(\mathbf{x}(t), \mathbf{u}(t)),
\end{equation} 
\noindent where $\tau$ is the system's response time and $\mathbf{x}(t)$ is the LA-VCSEL's state at time $t$ distributed across its surface.
State $\mathbf{x}(t)$ comprises the local optical field amplitude, the optical phase and the electronic carrier's divided between spin up and spin down populations \cite{Mulet2002}.
Equation (\ref{eq:LAVCSEL}) in its full expression is highly non-trivial, and interestingly a detailed numerical modelling of LA-VCSELs for long time-scales remains a highly challenging task due to their inherent nonlinearity and high dimensionality.
The ordinary differential Eq. (\ref{eq:LAVCSEL}) is structurally comparable to the concept of Neural Ordinary Differential Equations \cite{Chen2018}, and we leverage state $\mathbf{x}(t)$ to implement a recurrent neural network.
As we will operate $\mathbf{x}(t)$ an order of magnitude slower than any of its inherent time scales we only need to consider its steady state, and $\dot{\mathbf{x}}(t) = 0$.
Finally, the LA-VCSEL's physical variable used for computing is its optical emission power $\mathbf{P}(t)$.

For implementing programmable readout weights we image the LA-VCSEL's near-field onto DMDb (Vialux XGA 0.7" D4100), which applies Boolean output weights $\mathbf{W}^{\mathrm{out}}$ to the detected state variable $\mathbf{P}(t)$.
The weighted state is converted in output signal $y^{\mathrm{out}}$ via an optical detector (det):
\begin{equation}\label{eq:Output}
y^{\mathrm{out}}(t) \propto \mathbf{W}^{\mathrm{out}}\mathbf{P}(t) .
\end{equation}
\noindent We image DMDb on the detector, and hence optical fields of individual DMD-mirrors do not spatially overlap.
This makes the detected signal proportional to the weighted sum of powers.
If, however, imaging onto the detector mixes the optical fields impinging from individual mirrors, one has to modify Eq. (\ref{eq:Output}) for coherent detection.
Finally, we make $y^{\mathrm{out}}(t)$ unitless and normalized by subtracting its mean value and normalize by its standard deviation.

\begin{figure}[t]
\centering
\includegraphics[width=1\columnwidth]{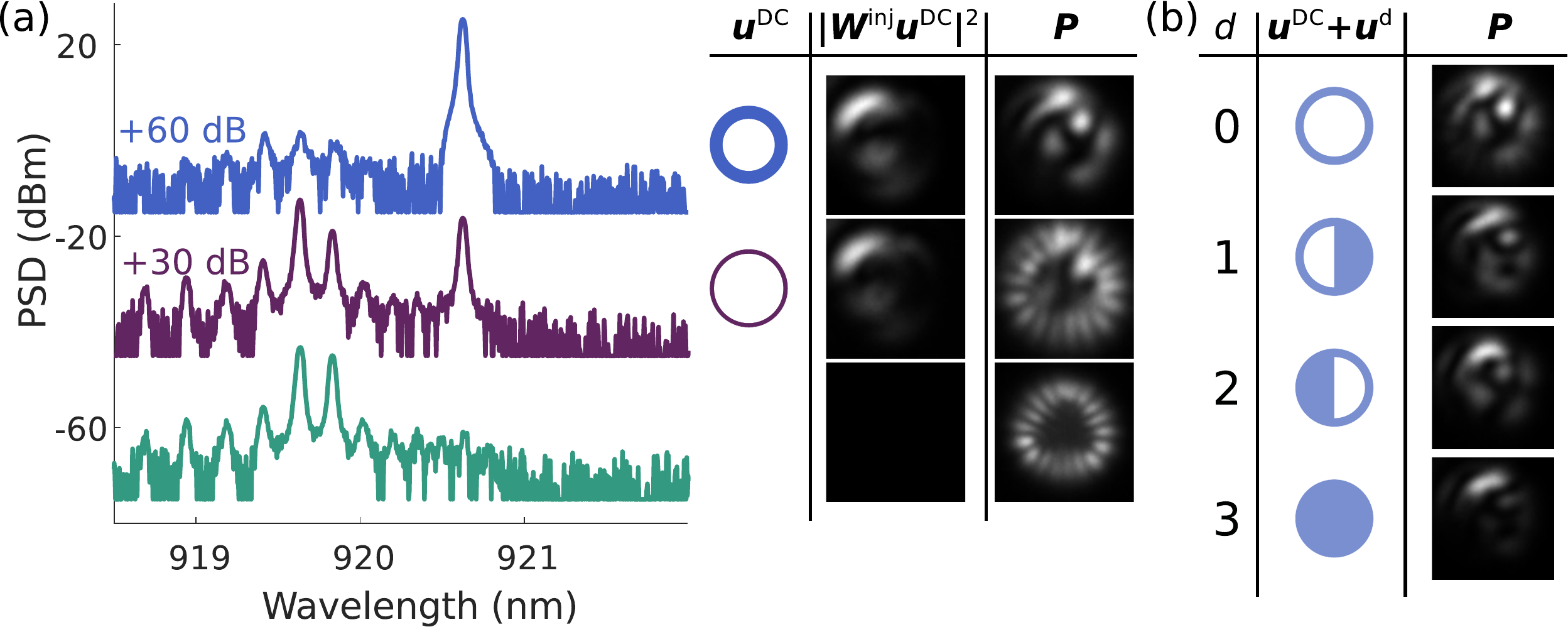}
\caption{ (a) Optical spectra, optical injection $\mathbf{W}^{\mathrm{inj}}\mathbf{u}^{\mathrm{DC}}$ and the LA-VCSEL's optical near field for the DC-locking ring $\mathbf{u}^{\mathrm{DC}}$.
	For clarity an offset of 30 dB (60 dB) was added to the middle (top) spectrum.
	(b) Total input information $\mathbf{u}^{\mathrm{in}}=\mathbf{u}^{\mathrm{DC}}+\mathbf{u}^{\mathrm{d}}$ for 2-bit encoding of number $d$, and LA-VCSEL response $\mathbf{x}$ to the corresponding optical injection.}
\label{fig:InjectionLock}
\end{figure}

Evolutionary learning will be used to optimize only $\mathbf{W}^{\mathrm{out}}$ \cite{Bueno2018}.
$\mathbf{W}^{\mathrm{in}}$ and $\mathbf{W}^{\mathrm{int}}$ do not partake in task-specific optimization, and we therefore implement a reservoir computer \cite{Jaeger2004}.
Output $y^{\mathrm{out}}(t)$ is recorded for a number of $t\in[1,\cdots,t^{\mathrm{T}}]$ discrete samples, i.e. our batch-size, for which we posses labeled target values $y^{\mathrm{T}}(t)$.
We compute mean square error $\epsilon_k = ||y_k^{\mathrm{out}}(t) - y^{\mathrm{T}}(t)||^2 \ / \ t^{T}$ at each learning epoch $k$, and at the transition from $k \rightarrow k+1$ the single weight $W^{\mathrm{out}}_{l(k),k}$ at position $l(k)$ is inverted.
If such weight inversion was beneficial ($\epsilon_{k+1}<\epsilon_{k}$) we keep this particular modification, if not we revert according to $W_{l(k+1),k+1}^{\mathrm{out}}=W_{l(k+1),k}^{\mathrm{out}}$.
In previous experiments on a spatial light modulator based setup \cite{Bueno2018} we found that for random selection of $l(k)$ error $\epsilon(k)$ converges exponentially on average \cite{Andreoli2020}, and we have analytically shown that this form of convergence is task independent \cite{Chretien2020}.

\section{\label{sec:Experiment} Experimental setup}

The experimental setup is schematically illustrated in Fig. \ref{fig:ConceptExperiment}(b).
The output of our fiber-coupled ECL is collimated via a 7.5~mm focal distance lens (Thorlabs AC050-008-B-ML).
The surface of DMDa is imaged with a magnification of 0.04 via two lenses (Thorlabs AC254-150-B-M and C110TMD-B) onto our 25~$\mu$m diameter, one meter long mm-fiber (Thorlabs M67L01).
DMDa displays input information $\mathbf{u}^{\mathrm{in}}$ with a frame rate of $\approx130$ frames per second, which currently sets the speed of our photonic ANN.
Using a 25~mm focal length lens (AC127-025-B-ML) in combination with a MAG10 near infrared optimized microscope objective (MO, Olympus LMPLN10XIR), the mm-fiber's near-field at its output is imaged on top of our LA-VCSEL with a magnification of 0.72.
The LA-VCSEL is imaged onto a CMOS camera (ids U3-3482LI-H) using a 100~mm focal distance lens (AC254-100-B-M).
Within the same optical path we placed a non-polarizing 50/50 beam-splitter (Thorlabs CCM1-BS014/M), and the reflected signal is collected by a single mode fiber connected to a optical spectrum analyzer (OSA, Yokagawa AQ6370D).
Additional non-polarizing 90/10 (Thorlabs BSX10R) and 50/50 (Thorlabs CCM1-BS014/M) beam-splitters further direct optical injection and the LA-VCSEL's emission.
Finally, we image the LA-VCSEL onto DMDb with a 100~mm focal length lens (Thorlabs AC254-100-B-M), resulting in a magnification of 5.6, and an additional lens (Thorlabs AC254-045-B-M) focuses the reflected signal on a detector (Thorlabs PM100A, S150C).

Our device wafers are produced by metal-organic vapor-phase epitaxy (MOVPE) on heavily n-doped (001) surface oriented GaAs substrates.
The epitaxial VCSEL structure consists of an n-doped GaAs buffer layer, followed by a Si-doped (n-doped) 37-period hybrid (two section) bottom distributed Bragg reflector(DBR), a half-wavelength ($\lambda/2$ optically thick) cavity, a C-doped (p-doped) 14.5-period top DBR, and heavily p-doped Al$_{0.1}$Ga$_{0.9}$As and GaAs current spreading layers each $\lambda/2$ optically thick.
The bottom n-doped DBR is composed of 33 periods of Al$_\mathrm{x}$Ga$_{1-\mathrm{x}}$As with x = 0.05 high refractive index layers and x = 1.0 low refractive index layers, followed by 4 periods of x = 0.05 and x = 0.92 layers. The top p-doped DBR is similar but with alternating x = 0.1 and x = 0.92 Al$_\mathrm{x}$Ga$_{1-\mathrm{x}}$As layers. All DBRs include linearly graded interfaces between the alternating low and high refractive index layers to reduce series resistance. 
The optical cavity consists of five compressively strained $\sim 4.2$~nm-thick InGaAs quantum wells (QWs) surrounded by six strain-compensating $\sim 5.1$~nm-thick GaAsP barrier layers that are in tension. From double crystal X-ray diffraction measurements on QW calibration structures the net active region strain is close to zero (as designed). 
The ensemble of QWs and barriers is centered in the $\lambda/2$ optical cavity and the remaining layers of the optical cavity (on both sides of the QWs and barrier layers) are Al$_\mathrm{x}$Ga$_{1-\mathrm{x}}$As layers stepped from x = 0.4 to x = 0.92, and then ending with 20 nm-thick (before oxidation) Al$_{0.98}$Ga$_{0.02}$As layers. The two x = 0.98 layers reside half (10 nm) within and half (10 nm) outside the optical cavity. 
The two x = 0.98 layers are selectively thermally oxidized during device processing to form wave-guiding and current confining oxide apertures with tapered ends (that minimize scattering losses). The estimated oxidation length (from the top mesa edge inward) is roughly 9-10 $\mu$m, based on a series of selective thermal oxidation experiments on cleaved slivers of the VCSEL material - yielding a plot of the oxidation length (in from a mesa edge) versus the oxidation time.
On resonance (in our mode simulations) the optical field intensity nodes that define the $\lambda/2$ optical cavity fall at the approximate center of the two x = 0.98 layers. We fabricate top-surface-emitting VCSELs onto quarter wafer pieces with high frequency co-planar ground-signal-ground (GSG) contact pads suitable for on-wafer testing via a standard GSG electrical probe head with pin to pin spacing of 150 $\mu$m using the fabrication methods reported in \cite{Haghighi2020}.

\section{\label{sec:InjStates}Optical injection and ANN states}

\begin{figure}[t]
\centering
\includegraphics[width=1\columnwidth]{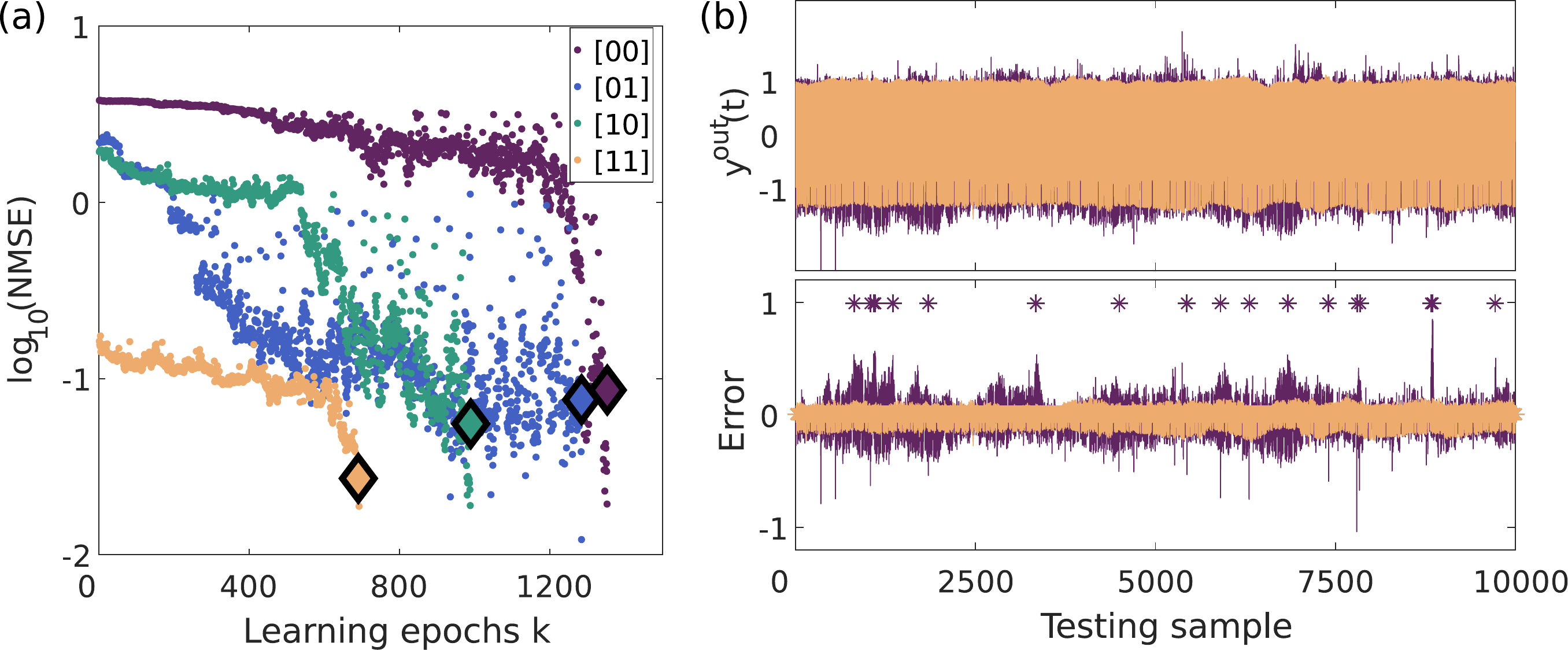}
\caption{ (a) Convergence on training set for classifying bits $<00>$, $<01>$, $<10>$ and $<11>$.
	Diamond symbols are the corresponding average testing NMSE.
	(b) Error on independent testing set with 10000 samples for classifying $<00>$ and $<11>$.
	The top panel shows $y^{\mathrm{out}}(t)$, and data for the lower injection power for digit $<00>$ exhibits slow undulations.
	The lower panel shows error $y^{\mathrm{out}}(t)-y^{\mathrm{T}}(t)$ as lines and the symbol classification error after thresholding as stars.}
\label{fig:Results}
\end{figure}

LA-VCSELs are highly complex photonic systems, and for emission apertures in excess of $\sim3~\mu$m their optical output field usually becomes multimode.
Optical injection into such devices has been extensively studied \cite{Ackemann1999,Ackemann2000}, and here we injection-lock such a device to a complex optical input field for the first time.
Figure \ref{fig:InjectionLock}(a) shows the device's optical response under DC injection using $\mathbf{u}^{\mathrm{DC}}$ with different rings.
The top, middle and bottom row were recorded using a ring with 15, 5 and 0 pixels thickness, respectively, which also is illustrated in the column labeled $\mathbf{u}^{\mathrm{DC}}$.
The bottom row therefore corresponds to the free-running laser without injection.
The table shows the configuration of DMDa (left column), the spatial distribution of the resulting optical injection $|\mathbf{W}^{\mathrm{inj}}\mathbf{u}^{\mathrm{DC}}|^2$ (center column) and the LA-VCSEL's spatial emission in response to the injected signal (right column).

We bias the laser with $I^{\mathrm{bias}}=9~$mA, which is slightly above its threshold ($I^{\mathrm{bias}}=1.28~I^{\mathrm{th}}$), where it emits a total of $P=1.68$~mW.
In an initial scan we tuned the injection laser's wavelength to the VCSEL's highest susceptibility to optical injection locking, which we found for injecting into a mode located at $\lambda^{\mathrm{inj}}=920.58~$nm.
For the free-running laser we found the expected multimode optical emission spectrum, and its optical emission profile shows the common set of high-order Laguerre-Gaussian modes \cite{Degen1999}.
A DC-lock ring width of 5 pixels results in injecting $P^{\mathrm{inj}}=0.05~$mW, for which the laser's emission spectrum as well as its spatial emission profile is already significantly perturbed.
Widening the DC-lock ring to 15 pixels increases the injection power to $P^{\mathrm{inj}}=0.7~$mW, for which the device is fully locked to the injected field.
The laser's free-running modes are suppressed by more than 30 dB, and its emission profile significantly differs from its free-running \emph{and} injected signal.
This already demonstrates the non-trivial and non-uniform transformation of optically injected patterns.
Figure \ref{fig:InjectionLock}(b) shows the resulting photonic ANN state for the example of injecting the 4 possible configurations $d\in[0,1,2,3]$ of 2-bit symbols.
Individual responses significantly differ for each case, which is a prerequisite to differentiate individual digits.

\section{\label{sec:learning} ANN performance of the LA-VCSEL}

As starting point for our photonic ANN's performance evaluation we trained a single output classifier to identify one of the four digits in a 2-bit header.
Training sequences of $t^{\mathrm{T}}=50$ samples were comprised to 50\% of the digit to be classified, and to 50\% of the remaining three digits.
The system was trained until the training error $\epsilon_{k}$ dropped below $\epsilon<2\cdot10^{-2}$, and in Fig. \ref{fig:Results}(a) we show the resulting convergence.

For all four tasks our system converged to below the targeted training error, however starting from strongly differing initial performances.
As reason we assume the injection conditions consequence of the different $\mathbf{u}^{\mathrm{in}}$ configurations, where $P^{\mathrm{inj}}$ is lowest for $d=0$ and highest for $d=3$.
Additionally, for the case of classifying bits $<00>$ the LA-VCSEL ANN needs to invert the power-dependency: the digit with the lowest injection needs to be transformed into the highest power in our output channel.
This observation is consistent with the initial classification errors for bits $<01>$ and $<10>$ starting from comparable levels, and with digit $<11>$ starting from the lowest level.

\begin{figure}[t]
\centering
\includegraphics[width=0.75\columnwidth]{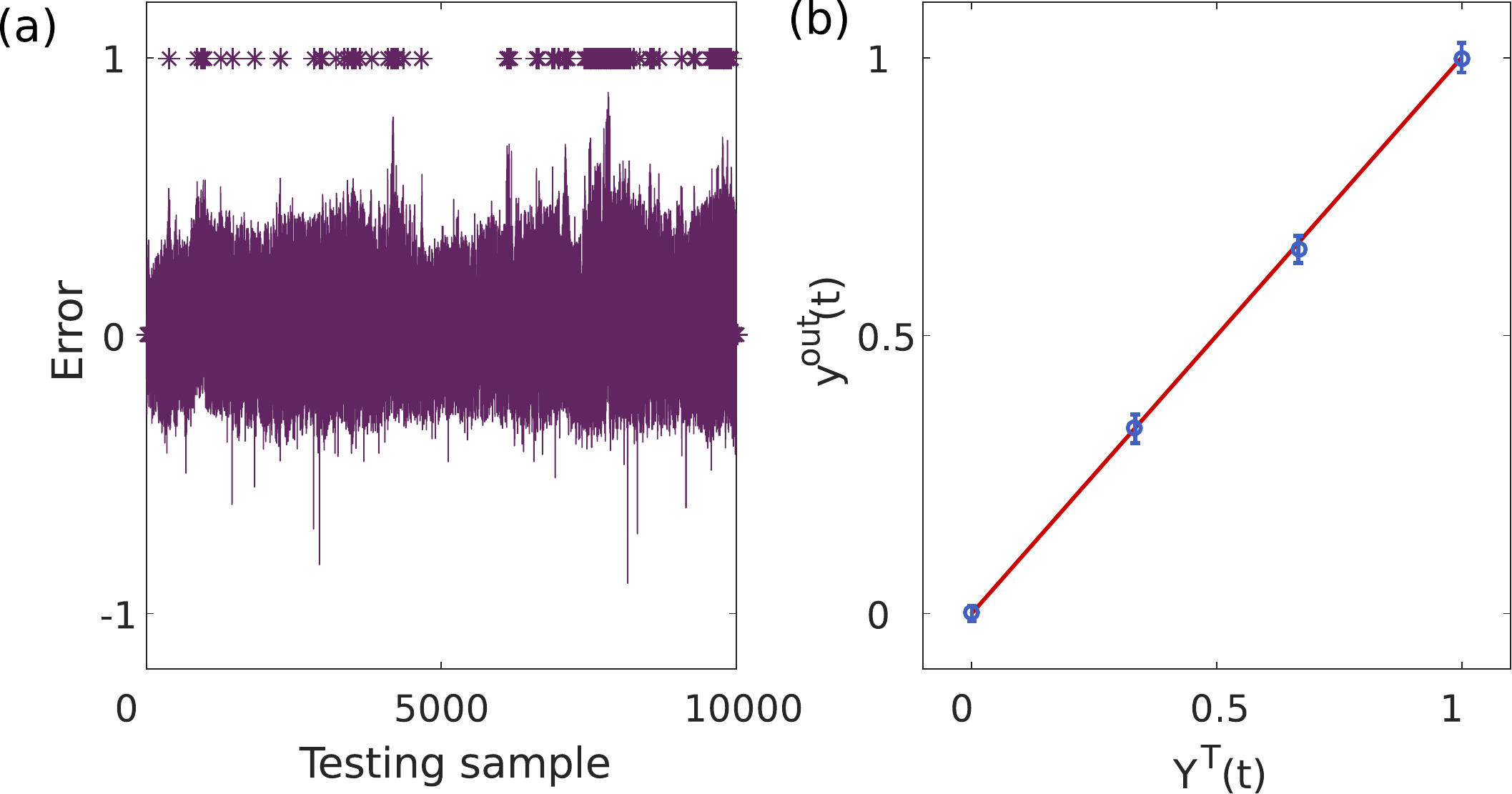}
\caption{ (a) Error on the XOR task for a testing set with 10000 samples.
	Data shows error $y^{\mathrm{out}}(t)-y^{\mathrm{T}}(t)$ as lines and the symbol classification error after thresholding as stars.
	(b) Performance in digital to analog conversion (DAC) for 10000 test samples, with the red line as the ideal case.}
\label{fig:ResultsXORDAC}
\end{figure}

After successful training, the system was tested with an independent test-sequence comprising 1000 samples, which for statistics we repeated 10 times.
Average testing errors correspond to the filled diamonds in Fig. \ref{fig:Results}(a).
Again, they reflect the stability of the different locking conditions.
Strongest injection (bits $<11>$) produces a testing error nearly identical to the training error, and lower injection powers result in stronger training vs. testing error divergence.
Figure \ref{fig:Results}(b) shows the temporally resolved testing sequences for bits $<00>$ and $<11>$.
Output $y^{\mathrm{out}}(t)$ for both cases is shown in the upper panel, and the stronger undulations for the case of classifying bits $<00>$ are clearly visible.
The lower panel shows error $y^{\mathrm{out}}(t)-y^{\mathrm{T}}(t)$ as lines, while the stars show a classification performance after threshold, i.e. $(y^{\mathrm{out}}(t)>0.5)-y^{\mathrm{T}}(t)$.
Therefore, stars in Fig. \ref{fig:Results}(b) located at amplitude 1 are samples for which our photonic ANN misclassified the corresponding digit. 
The number of misclassified samples normalized by the test sequence's length results in the symbol error rate (SER).
For the 10000 testing samples we obtained SER=($2.5\cdot10^{-3}$, $<1\cdot10^{-4}$, $1.1\cdot10^{-3}$, $<1\cdot10^{-4}$) for bits $<00>$, $<01>$, $<10>$, $<11>$, respectively.
On average we therefore obtain SER$<0.9\cdot10^{-3}$.

As a final test on the 2-bit task we train the LA-VCSEL ANN to: (i) implement an XOR logic operation; and (ii) to implement a DAC.
The XOR task is a classical benchmark to demonstrate that an ANN is capable to implement nonlinear transformation.
As such it demonstrated that single layer perceptrons \cite{Rosenblatt1958} failed in this tasks \cite{Minsky1969}, which ultimately motivated the development of deeper neural networks.
DAC places a similar, yet slightly easier challenge to an ANN.
Figure \ref{fig:ResultsXORDAC}(a) shows the error of the XOR task, illustrating $y^{\mathrm{out}}(t)-y^{\mathrm{T}}(t)$ as a line and as previously the misclassifications after thresholding as stars.
The obtained SER for the XOR task is 2.9$\cdot10^{-2}$, again after 10000 testing points.
Finally, the performance in DAC is shown in Fig. \ref{fig:ResultsXORDAC}(b), with the red line as the ideal case.
Here we find that the system excellently approximates the output target with a standard deviation of 5.4$\cdot10^{-2}$.

\section{\label{sec:Conclusions} Conclusions}

We demonstrated a spatially-extended ANN based on an injection locked LA-VCSEL with more than 130 artificial photonic neurons.
All ANN connections are implemented in photonic hardware, and the network's operation is fully parallel. 
Trainable readout connections are realized using a DMD, while static input connections are obtained via the complex transfer matrix of a mm-fiber.
Finally, the information input was provided via another DMD.

We trained and tested our photonic ANN's performance against tasks involving sequences of 2-bits, i.e. pattern classification, the XOR task and DAC.
Our LA-VCSEL based ANN shows excellent performance in all, achieving long-term stability of the trained states during the much longer testing sequences.
Finally, our demonstration is only a proof-of-concept.
The ANN's present operation speed is orders of magnitude below its potential bandwidth, and the speed of convergence can most likely be significantly improved with more advanced training concepts.

Our photonic ANN's future potential relies on the fact that the entire system is capable of fully parallel operation in the multi-GHz range.
We therefore project a potential speed-up by around 9 orders of magnitude.
Noteworthy, such bandwidths enable direct applications in ultra-fast systems, for example to implement real time closed loop control for novel ultra-fast light sources \cite{Genty2020}.
As all weights implemented in our system are passive or do not require fast modulation, the dynamical energy consumption of our photonic ANN is primarily by injection laser and the LA-VCSEL.
Considering the here reported optical powers as well as common semiconductor laser wall plug efficiencies, large scale and ultra-fast photonic ANNs consuming less than 1 W globally are realistic.

\section{Acknowledgment}

The authors acknowledge the support of the Region Bourgogne Franche-Comté.
This work is supported by the EUR EIPHI program (Contract No. ANR-17-EURE- 0002), by the Volkwagen Foundation (NeuroQNet I\&II), by the French Investissements d’Avenir program, project ISITE-BFC (contract ANR-15-IDEX-03), by the German Research Foundation
(Deutsche Forschungsgemeinschaft - DFG) via the Collaborative Research Center (Sonderforschungsbereich - SFB) 787, by the European Union’s Horizon 2020 research and innovation programme under the Marie Skłodowska-Curie grant agreements No. 860830 (POST DIGITAL). Also, author X.P. receives funding from the Marie Skłodowska-Curie grant agreement No. 713694 (MULTIPLY).

\bibliographystyle{ieeetr}
\bibliography{bibliography}

\end{document}